\documentclass[sigconf]{acmart}
\usepackage{hyperref}
\usepackage{url}
\usepackage{amsmath}
\usepackage{graphicx}
\usepackage{amsthm}
\usepackage{cite}
\usepackage{listings}
\usepackage{algorithm}
\usepackage{algpseudocode}
\usepackage{booktabs}
\usepackage{hyperref}

\newtheorem{theorem}{Theorem}[section] 
\newtheorem{definition}[theorem]{Definition} 
\newtheorem{proposition}[theorem]{Proposition} 

\newcommand{\hyperboloid}{\mathcal{H}} 
\newcommand{\logmap}[2]{\log_{#1}^{#2}} 
\newcommand{\expmap}[2]{\exp_{#1}^{#2}} 
\newcommand{\origin}{o} 
\newcommand{\Wop}{\mathbf{W} \otimes_c^\mathcal{L}} 

\theoremstyle{definition}
\AtBeginDocument{%
  }

\acmConference[ACM SIGMOD 2025 GRADES NDA Workshop]{International Conference on Management of Data}{June 22--27,
  2025}{Berlin, Germany}
  \acmYear{2025}

\renewcommand\footnotetextcopyrightpermission[1]{} 
\begin{document}

\title[]{Can we ease the Injectivity Bottleneck on Lorentzian Manifolds for Graph Neural Networks?}

\author{Srinitish Srinivasan}
\email{srinitish.srinivasan2021@vitstudent.ac.in}
\affiliation{%
  \institution{Vellore Institute of Technology}
  \country{India}
}

\author{Omkumar CU}
\email{omkumar.cu@vit.ac.in}
\affiliation{%
  \institution{Vellore Institute of Technology}
  \country{India}
}

\renewcommand{\shortauthors}{Srinitish et al.}


\begin{abstract}
While hyperbolic GNNs show promise for hierarchical data, they often have limited discriminative power compared to Euclidean counterparts or the WL test, due to non-injective aggregation. To address this expressivity gap, we propose the Lorentzian Graph Isomorphic Network (LGIN), a novel HGNN designed for enhanced discrimination within the Lorentzian model. LGIN introduces a new update rule that preserves the Lorentzian metric while effectively capturing richer structural information. This marks a significant step towards more expressive GNNs on Riemannian manifolds. Extensive evaluations across nine benchmark datasets demonstrate LGIN's superior performance, consistently outperforming or matching state-of-the-art hyperbolic and Euclidean baselines, showcasing its ability to capture complex graph structures. LGIN is the first to adapt principles of powerful, highly discriminative GNN architectures to a Riemannian manifold. The code for our paper can be found at https://github.com/Deceptrax123/LGIN
\end{abstract}


\received[received]{4 April 2025}
\received[accepted]{1 May 2025}

\maketitle

\section{Introduction}
Graphs model complex relationships, but standard Euclidean Graph Neural Networks (GNNs) (\citep{kipf2016semi},\citep{velivckovic2017graph}) struggle with graphs exhibiting strong hierarchical or multi-relational structures. Hyperbolic geometry, with its exponential volume growth, is better suited for embedding such data with minimal distortion \citep{zhu2023lorentzian}. In parallel, Graph Isomorphic Networks (GINs) \citep{xu2018powerful} achieve high expressivity in Euclidean space, equivalent to the 1-dimensional Weisfeiler-Lehman test, through injective aggregation. However, extending GIN's expressivity to hyperbolic space is challenging due to the need to preserve geometric properties and metric constraints during message passing.

Motivated to bridge this gap, we propose the Lorentzian Graph Isomorphic Network (LGIN), the first attempt at an isomorphic GNN incorporating constant curvature. LGIN employs a novel hyperbolic-tangent-hyperbolic approach to address metric preservation. To approximate the discriminative power of Euclidean GINs, LGIN introduces a modified cardinality-scaled aggregation rule, preserving multiset information often lost in hyperbolic settings. This design yields a GNN with expressivity at least approximately equivalent to the 1-dimensional Weisfeiler-Lehman test. Empirical results demonstrate LGIN's superior discriminative capabilities and performance compared to existing hyperbolic GNNs like LGCN across various benchmarks. Therefore, we summarize our key contributions as follows:
\begin{itemize}
    \item To the best of our knowledge, \textbf{this is the first study that makes an attempt to study an isomorphic Graph Neural Network(GNN)} that incorporates both constant and variable curvature information.
    \item We show that GNNs built using the\textbf{ hyperbolic-tangent-hyperbolic approach while preserving the metric tensor at the origin has high discriminative capability}.
    \item We propose the theoretical framework for \textbf{Lorentzian Graph Isomorphic Netork(LGIN)}, with\textbf{ a modified cardinality-scaled aggregation rule to preserve the cardinality information of multisets and a new update rule}, as an approximately powerful alternative to Graph Isomorphic Network(GIN) and other state of the art hyperbolic graph neural networks such as LGCN.
    \item We conduct several experiments across 9 benchmark datasets ranging from MoleculeNet to TU benchmarks and show that our approach consistently outperforms or matches most baselines.
\end{itemize}
\section{Related Work}

In this section, we outline related work concerning Hyperbolic Graph Neural Networks (HGNNs) and Graph Expressiveness. Hyperbolic Graph Neural Networks have shown promise for modeling complex and hierarchical structures, benefiting from hyperbolic space's natural ability to embed such data with low distortion \citep{liu2019hyperbolic}. Various approaches have been explored, including manifold-agnostic generalizations \citep{liu2019hyperbolic}, utilizing specific models like the Poincare disk with normalization layers \citep{khatir2022unification}, adapting curvature based on data \citep{fu2023adaptive}, and developing methods like Lorentzian Graph Convolutional Networks (LGCN) which use centroid aggregation in the Lorentz model \citep{zhang2021lorentzian}. HGNNs have found successful applications in diverse domains, including brain connectivity, drug discovery, and knowledge graphs.

Understanding the expressive power of GNNs is crucial for distinguishing graph structures. The 1-dimensional Weisfeiler-Leman (WL) test provides a benchmark for graph isomorphism testing. Graph Isomorphism Networks (GINs) \citep{xu2018powerful} represent a class of message-passing GNNs in Euclidean space that are provably as powerful as the 1-WL test, primarily due to their use of injective aggregation functions. This highlights the importance of aggregation mechanisms for achieving high discriminative capabilities. We have elaborated on the related work in appendix \ref{app:related}.

\section{Proposed Framework}
In this section, we discuss the framework for an approximately powerful GNN on the Lorentzian model. Note that this can be extended to other hyperbolic models as well through their respective exponential and logarithmic maps. We discuss the properties of the manifold that allow for discriminative GNNs, however such frameworks are limited by the need of preserving the metric tensor and attention aggregated weights. Throughout our study, we consider more importance to metric tensor preservation than injectivity of aggregation. 
\subsection{Overview}\label{sec:overview}
\begin{theorem}[\textbf{Manifold Property for powerful GNNs}] \label{theo:iso}
Let $G_1 = (V_1, E_1)$ and $G_2 = (V_2, E_2)$ be two graphs. Consider a smooth Lorentzian manifold $(\mathcal{M}, g)$ of constant negative curvature $c < 0$, where $g$ is the Lorentzian metric tensor. Suppose there exists an embedding

\begin{equation} \label{eq:embedding_phi_refined}
\Phi: \mathcal{G} \to \mathcal{M}
\end{equation}
that maps vertices of any graph $G \in \mathcal{G}$ into the hyperboloid model $\mathbb{H}^n$ and preserves graph structural properties (e.g., graph distance) through the Lorentzian distance on $\mathcal{M}$. If $G_1$ and $G_2$ are non-isomorphic, then their embeddings $\Phi(G_1)$ and $\Phi(G_2)$ cannot be related by a global isometry $f$ of $(\mathcal{M}, g)$, i.e., there does not exist a diffeomorphism
\begin{equation} \label{eq:diffeomorphism_f_refined}
f : \mathcal{M} \to \mathcal{M}
\end{equation}
such that
\begin{equation} \label{eq:isometry_condition_refined}
f^*g = g \quad \text{and} \quad f \circ \Phi(G_1) = \Phi(G_2)
\end{equation}
\end{theorem}
We present the proof for theorem \ref{theo:iso} in appendix \ref{app:iso}. Therefore, we can demonstrate that a graph neural network mapping node features onto a hyperboloid can achieve representational power approximately comparable to the Weisfeiler-Lehman (WL) test owing to the property of the metric embeddings on a hyperbolic manifold, thus describing a desirable embedding. However, the following must be taken into consideration:
\begin{itemize}
    \item  A powerful graph neural networks requires strictly injective aggregation and readout functions\citep{xu2018powerful}. However, directly translating strictly injective operations to Riemannian manifolds while preserving geometric properties is non-trivial. Our goal is to design a network that achieves high expressive power by carefully approximating these properties within the Lorentzian geometry.
    \item Additionally, geometric consistency i.e the preservation of metric tensors of the hyperboloid model needs to be ensured.
\end{itemize}
To account for the above considerations, we have provided a detailed explanation in the following sections. 

\subsection{Injectivity of Lorentz Transformation}\label{sec:injectivity}
We show the injectivity of the Lorentz Transformation on the manifold and through moment injectivity proposed by \citep{amir2023neural}. To ensure the injectivity of the Lorentz transformation, we follow the hyperbolic-tangential-hyperbolic approach \citep{zhang2021lorentzian}. This is because the tangent space at a point is locally isometric to Euclidean space. The Lorentz transformation used in our proposed method is given by:
\begin{equation}
    \mathbf{W} \otimes_c^\mathcal{L} \mathbf{x}^\mathcal{L} := \exp_o^c \left( 0, \mathbf{W} \log_o^c \left( \mathbf{x}^\mathcal{L} \right)_{[1:n]} \right),
    \label{eq:lorentz}
\end{equation}
where $\mathbf{x}^\mathcal{L} \in \mathcal{L}_c^n$, $\mathbf{W} \in \mathbb{R}^{d \times n}$. This method ensures the first coordinate is consistently zero, signifying that the resultant transformation is invariably within the tangent space at $\mathbf{o}$. In the following sections. we prove the injectivity of the Lorentz transformation.

\subsubsection{Injectivity on the Manifold}

\begin{proposition} 
\label{prop:full_rank}
The Lorentzian transformation $\Wop: \hyperboloid^n \to \hyperboloid^d$ defined by
\[ \Wop x^\mathcal{L} = \expmap{\origin}{c} \left(0, \mathbf{W} \left(\logmap{\origin}{c} x^\mathcal{L}\right)_{[1:n]}\right) \]
is injective if and only if the weight matrix $\mathbf{W} \in \mathbb{R}^{d \times n}$ has full column rank, i.e., $\text{rank}(\mathbf{W}) = n$. This requires $d \ge n$.
\end{proposition}
The proof of the above proposition is detailed in appendix \ref{app:full_rank}. Based on proposition \ref{prop:full_rank}, we ensure that the transformation matrix $\mathbf{W}$ has a full column rank by setting the number of output dimensions to be greater than the number of input dimensions for each layer. 
\begin{table*}[t!]
    \centering
    \setlength{\tabcolsep}{2pt} 
    \caption{\textbf{Performance of GNN Variants against LGIN}. The best score for all variants are marked in \textbf{bold} along with LGINs comparable with GIN variants based on a paired t-test with significance level of 10\%. Models performing significantly better are marked in \textbf{bold} with an asterisk. Evaluation standard defined as per \citep{xu2018powerful}}
    \begin{tabular}{lcccc}
    \toprule
    \textbf{Datasets} & \textbf{MUTAG} & \textbf{PROTEINS} & \textbf{PTC} & \textbf{NCI1} \\
    \hline
    \# graphs & 188 & 1113 & 344 & 4110 \\
    \# classes & 2 & 2 & 2 & 2 \\
    Avg \# nodes & 17.9 & 39.1 & 25.5 & 29.8 \\
    \hline
    WL subtree & 90.4 $\pm$ 5.7 & 75.0 $\pm$ 3.1 & 59.9 $\pm$ 4.3 & \textbf{86.0 $\pm$ 1.8\text{*}} \\
    DCNN & 67.0 & 61.3 & 56.6 & 62.6 \\
    PATCHYSAN & \textbf{92.6 $\pm$ 4.2\text{*}} & 75.9 $\pm$ 2.8 & 60.0 $\pm$ 4.8 & 78.6 $\pm$ 1.9 \\
    DGCNN & 85.8 & 75.5 & 58.6 & 74.4 \\
    AWL & 87.9 $\pm$ 9.8 & - & - & - \\
    \hline
    SUM-MLP (GIN-0) & 89.4 $\pm$ 5.6 & 76.2 $\pm$ 2.8 & 64.6 $\pm$ 7.0 & 82.7 $\pm$ 1.7 \\
    SUM-MLP (GIN-$\epsilon$) & 89.0 $\pm$ 6.0 & 75.9 $\pm$ 3.8 & 63.7 $\pm$ 8.2 & 82.7 $\pm$ 1.6 \\
    SUM-1-LAYER & 90.0 $\pm$ 8.8 & 76.2 $\pm$ 2.6 & 63.1 $\pm$ 5.7 & 82.0 $\pm$ 1.5 \\
    MEAN-MLP & 83.5 $\pm$ 6.3 & 75.5 $\pm$ 3.4 & 66.6 $\pm$ 6.9 & 80.9 $\pm$ 1.8 \\
    MEAN-1-LAYER (GCN) & 85.6 $\pm$ 5.8 & 76.0 $\pm$ 3.2 & 64.2 $\pm$ 4.3 & 80.2 $\pm$ 2.0 \\
    MAX-MLP & 84.0 $\pm$ 6.1 & 76.0 $\pm$ 3.2 & 64.6 $\pm$ 10.2 & 77.8 $\pm$ 1.3 \\
    MAX-1-LAYER (GraphSAGE) & 85.1 $\pm$ 7.6 & 75.9 $\pm$ 3.2 & 63.9 $\pm$ 7.7 & 77.7 $\pm$ 1.5 \\
    \hline
    SUM-Lorentz (LGIN Variable Curvature) & \textbf{88.1 $\pm$ 5.0} & \textbf{79.1 $\pm$ 4.4\text{*}} & \textbf{68.2 $\pm$ 5.1\text{*} }& \textbf{82.3 $\pm$ 1.5} \\
    SUM-Lorentz (LGIN Fixed Curvature) & \textbf{90.6 $\pm$ 3.9} & \textbf{76.9 $\pm$ 2.5 }& \textbf{68.4 $\pm$ 5.2\text{*}} & \textbf{83.2 $\pm$ 1.6} \\
    \bottomrule
    \end{tabular}
\label{tab:gin_base}
\end{table*}
\subsubsection{Moment Injectivity}
 From \citep{amir2023neural}, we have the following result for the Euclidean space:
Consider shallow neural networks \( f : \mathbb{R}^d \rightarrow \mathbb{R}^m \) of the form
\begin{equation}
    f(\mathbf{x}; A, \mathbf{b}) = \sigma(A\mathbf{x} + \mathbf{b}), \quad A \in \mathbb{R}^{m \times d}, \ \mathbf{b} \in \mathbb{R}^m,
\end{equation}
with the activation function \( \sigma : \mathbb{R} \rightarrow \mathbb{R} \) applied entrywise to \( A\mathbf{x} + \mathbf{b} \). Suppose that \( \sigma \) is analytic and non-polynomial; such activations include the sigmoid, softplus, tanh, swish, and sin. For a large enough \( m \), such networks \( f(\mathbf{x}; A, \mathbf{b}) \) with random parameters \( A, \mathbf{b} \) are moment-injective on \( \mathcal{M}_{\leq n}(\Omega) \) and on \( \mathcal{S}_{\leq n}(\Omega) \); namely, their induced moment functions \( \hat{f} \) are injective. This holds for various natural choices of \( \Omega \). \( \hat{f} \) is given by \( f : \Omega \rightarrow \mathbb{R}^m \) that induces a moment function 
\[
\hat{f} : \mathcal{M}_{\leq n}(\Omega) \rightarrow \mathbb{R}^m
\]
defined by
\begin{equation}
    \hat{f}(\mu) = \int_{\Omega} f(x) d\mu(x) = \sum_{i=1}^n w_i f(x_i), \quad \text{where} \quad \mu = \sum_{i=1}^n w_i \delta_{x_i}.
    \label{eq:moment}
\end{equation}
Now, extending the theory from eq.\ref{eq:moment} to the hyperboloid, we have the transformation defined in eq.\ref{eq:lorentz}, where $x^\mathcal{L}$ is first mapped to the tangent space. We already know that the tangent space at a point is locally isometric to the Euclidean space. Therefore, by the isometry, the moment function in this transformed space becomes:
\begin{equation}
    \hat{f}(\mu) = \int_{\mathcal{L}_c} f(\mathbf{x}^{\mathcal{L}}) d\mu(\mathbf{x}^{\mathcal{L}}) = \sum_{i=1}^n w_i f(\mathbf{x}_i^{\mathcal{L}}), \quad \mu = \sum_{i=1}^n w_i \delta_{\mathbf{x}_i^{\mathcal{L}}}.
\end{equation}
where $x^{\mathcal{L}}$ resides in the tangent space. Therefore,we use a shallow(2-Layer) Lorentzian network based on transformation equation \ref{eq:lorentz} with a pointwise Tanh/Sigmoid function and ensuring $\mathbf{W}$ has a full column rank. 

\subsection{Lorentzian Graph Isomorphic Network}

Based on the conditions mentioned in section \ref{sec:overview}, proof in 
 section \ref{sec:injectivity} and corollary 6 in \citep{xu2018powerful}, we finally define the update rule for the Lorentzian Graph Isomorphic Network(LGIN) as follows:
 \begin{equation}
     \mathbf{x}^{\mathcal{L}^k}=\text{LT}^k(\exp_o^c((1+\epsilon^k).\mathcal{P}_{x^{{L}^{k-1}}\longrightarrow T}(\log_o^c(\mathbf{x}^{\mathcal{L}^{k-1}}))+\log_o^c(T(\mathbf{x}^{\mathcal{L}^{k-1}}))))
     \label{eq:update}
 \end{equation}
 where $\mathbf{x}^\mathcal{L}$ is a point in the Lorentzian space, $\text{LT}$ represents the Lorentz transformation which is a 2-layer model with Tanh activation and output features greater than input, $k$ is the step and $T$ is the output of neighbor aggregation given by the cardinality aware Lorentz centroid method as follows:
\begin{equation}
T(\mathbf{x}_i^{\mathcal{L}^{k-1}}) := \frac{\sqrt{c}\sum_{j \in \mathcal{N}_i} \alpha_{ij} \mathbf{x}_j^{\mathcal{L}}}{(1+|\mathcal{N}_i|) \lVert \sum_{j \in \mathcal{N}_i} \alpha_{ij} \mathbf{x}_j^{\mathcal{L}} \rVert_{\mathcal{L}}}
\label{eq:aggregate}
\end{equation}
where $\alpha_{ij}$ refers to weights with squared hyperbolic distance to lay emphasis on geometry. It is given by:
\begin{equation}
\alpha_{ij} = \text{softmax}_{j \in \mathcal{N}(i)} \left( -d_c^2(\mathbf{x}_i^{\mathcal{L}}, \mathbf{x}_j^{\mathcal{L}}) \right)
\end{equation}

\subsubsection{Justification of Lorentz Centroid}
Equation \ref{eq:aggregate} ensures embeddings reside on the hyperbolic manifold after aggregation i.e the preservation of metric tensors. Normalization prevents the node embeddings from collapsing into the origin or drifting too far. However, since attention-based aggregators are  ot strictly injective, it may lead to non-isomorphic graphs having the same embeddings. To mitigate this, we modify the Lorentz aggregation function to preserve the cardinality information of multisets during aggregation which can approximate the injective property\citep{zhang2020improving}. 

\subsubsection{Justification of Parallel Transport}
Parallel Transport in equation \ref{eq:update} preserves the hyperbolic structure and avoids distorting embeddings. The embeddings of two neighboring nodes reside in different tangent spaces because each node has its own local geometry due to curvature. A direct aggregation of features between nodes would lead to the mixing of incompatible spaces, thereby leading to node representations that are not completely representative of that node. We provide an empirical analysis by comparing the model's performance with and without parallel transport in the following section. 

\section{Experiments} 
In this section, we discuss the experimental conditions and results obtained from testing the Lorentzian Graph Isomorphic Network over various datasets. 
\subsection{Evaluations}
\subsubsection{Evaluations against Graph Isomorphic Network}

\begin{table}[h!]
    \centering
    \caption{\textbf{Detailed Analysis on Proteins, Enzymes and DD.} Test accuracy $\pm$ standard deviation. The best score is in \textbf{bold} and the second best is \underline{underlined}. Hyphen indicates scores not reported. Hyperbolic GNN baselines are marked with an asterisk(*).}
    \label{tab:ind}
    \small 
    \begin{tabular}{|c|c|c|c|}
        \hline
        \textbf{Model} & $L$ & \textbf{Test Acc.} $\pm$ \textbf{s.d.} & \textbf{Train Acc.} $\pm$ \textbf{s.d.} \\
        \hline
        \multicolumn{4}{|c|}{\textbf{ENZYMES}} \\
        \hline
        MLP & 4 & 55.833$\pm$3.516 & 93.062$\pm$7.551 \\
        \textit{vanilla} GCN & 4 & 65.833$\pm$4.610 & 97.688$\pm$3.064 \\
        GraphSage & 4 & 65.000$\pm$4.944 & 100.000$\pm$0.000 \\
        MoNet & 4 & 63.000$\pm$8.090 & 95.229$\pm$5.864 \\
        GAT & 4 & \underline{68.500$\pm$5.241} & 100.000$\pm$0.000 \\
        GatedGCN & 4 & 65.667$\pm$4.899 & 99.979$\pm$0.062 \\
        GIN & 4 & 65.333$\pm$6.823 & 100.000$\pm$0.000 \\
        RingGNN & 2 & 18.667$\pm$1.795 & 20.104$\pm$2.166 \\
        3WLGNN & 3 & 61.000$\pm$6.799 & 98.875$\pm$1.571 \\
        HGNN* & - & 51.300$\pm$6.100 & - \\
        H2H-GCN*& - & 61.300$\pm$4.900 & - \\
        LGIN Variable $c$(Ours) & 3 & 67.380$\pm$4.100 & 100.000$\pm$0.000 \\
        LGIN Fixed $c$(Ours) & 3 & \textbf{71.500$\pm$1.600} & 100.000$\pm$0.000 \\
        \hline
        \multicolumn{4}{|c|}{\textbf{DD}} \\
        \hline
        MLP & 4 & 72.239$\pm$3.854 & 73.816$\pm$1.015 \\
        \textit{vanilla} GCN & 4 & 72.758$\pm$4.083 & 100.000$\pm$0.000 \\
        GraphSage & 4 & 73.433$\pm$3.429 & 75.289$\pm$2.419 \\
        MoNet & 4 & 71.736$\pm$3.365 & 81.003$\pm$2.593 \\
        GAT & 4 & \underline{75.900$\pm$3.824} & 95.851$\pm$2.575 \\
        GatedGCN & 4 & 72.918$\pm$2.090 & 82.796$\pm$2.242 \\
        GIN & 4 & 71.910$\pm$3.873 & 99.851$\pm$0.136 \\
        RingGNN & 2 & - & - \\
        3WLGNN & 3 & - & - \\
        HGNN* & - & 75.800$\pm$3.300 & - \\
        H2H-GCN*  & - & \textbf{78.200$\pm$3.300} & - \\
        LGIN Variable $c$(Ours) & 3 & 73.730$\pm$3.200 & 100.000$\pm$0.000 \\
        LGIN Fixed $c$(Ours) & 3 & 70.300$\pm$3.500 & 100.000$\pm$0.000 \\
        \hline
        \multicolumn{4}{|c|}{\textbf{PROTEINS}} \\
        \hline
        MLP & 4 & 75.644$\pm$2.681 & 79.847$\pm$1.551 \\
        \textit{vanilla} GCN & 4 & 76.098$\pm$2.406 & 81.387$\pm$2.451 \\
        GraphSage & 4 & 75.289$\pm$2.419 & 85.182$\pm$3.489 \\
        MoNet & 4 & 76.452$\pm$2.898 & 78.206$\pm$0.548 \\
        GAT & 4 & 76.277$\pm$2.410 & 83.186$\pm$2.000 \\
        GatedGCN & 4 & 76.363$\pm$2.904 & 79.471$\pm$0.695 \\
        GIN & 4 & 74.117$\pm$3.357 & 75.351$\pm$1.267 \\
        RingGNN & 2 & 67.564$\pm$7.551 & 67.564$\pm$7.551 \\
        3WLGNN & 3 & 61.712$\pm$4.859 & 62.427$\pm$4.548 \\
        HGNN* & - & 73.700$\pm$2.300 & - \\
        H2H-GCN* & - & 74.400$\pm$3.000 & - \\
        LGIN Variable $c$(Ours) & 3 & \textbf{79.100$\pm$4.400} & 81.750$\pm$10.030 \\
        LGIN Fixed $c$(Ours) & 3 & \underline{76.900$\pm$2.500} & 84.180$\pm$5.530 \\
        \hline
    \end{tabular}
\end{table}
Table \ref{tab:gin_base} presents a performance comparison of LGIN variants against powerful Euclidean baselines, including GIN variants, WL kernels, and others, across MUTAG, PROTEINS, PTC, and NCI1 datasets. The proposed LGIN variants demonstrate competitive and often superior performance. Specifically, \textbf{LGIN with Fixed Curvature} achieves the highest accuracy on the \textbf{MUTAG} (90.6 $\pm$ 3.9) and \textbf{PTC} (68.4 $\pm$ 5.2) datasets. \textbf{LGIN with Variable Curvature} attains the best performance on the \textbf{PROTEINS} dataset (79.1 $\pm$ 4.4), showing statistically significant improvement ($p < 0.1$). While the WL subtree kernel leads on NCI1 (86.0 $\pm$ 1.8) and PATCHYSAN is competitive on MUTAG (92.6 $\pm$ 4.2), LGIN variants consistently rank among the top performers. These results underscore LGIN's ability to effectively leverage Lorentzian geometry and incorporated curvature for enhanced representation learning and discrimination across graphs with varying structures, demonstrating the benefits of adapting isomorphic principles to non-Euclidean spaces.

\subsubsection{Performance on Enzymes, Proteins and DD against state of the art graph  neural networks}
Table \ref{tab:ind} details performance across ENZYMES, DD, and PROTEINS, evaluating LGIN variants, Euclidean GNNs, and other HGNNs.
Performance on ENZYMES, DD, and PROTEINS is shown in Table \ref{tab:ind}. On ENZYMES, our LGIN Fixed model achieves the highest accuracy (71.5\%), significantly outperforming hyperbolic baselines, with GAT second. For DD, H2H-GCN leads (78.2\%), followed by GAT and HGNN; our LGIN Variable (73.70\%) is competitive. In PROTEINS, our LGIN Variable attains the highest accuracy (79.1\%), surpassing LGIN Fixed and other hyperbolic baselines. Overall, LGIN variants are highly competitive across these datasets. LGIN Fixed is best on ENZYMES, LGIN Variable on PROTEINS. While H2H-GCN is best on DD, LGIN variants perform strongly, highlighting their effectiveness on complex graph structures.

\subsubsection{Performance on MoleculeNet Datasets}
\begin{table}[h!]
\centering
\caption{\textbf{Analysis of our framework's performance on MoleculeNet baselines.} The best score is marked in \textbf{bold} and the second best score is \underline{underlined}. GIN models have been marked with an asterisk to indicate that our proposed model significantly outperforms GIN baselines. Hyphen indicates OOM on an 8GB M2 Macbook Pro. }
\label{tab:molnet}
\begin{tabular}{lccc}
\toprule
\textbf{Datasets}   & \textbf{BACE} & \textbf{BBBP} & \textbf{HIV} \\
\#Molecule          & 1513          & 2039          & 41127        \\
\#Tasks             & 1             & 1             & 1            \\
\hline
GAT                 & 83.60         & 65.48         & 69.26  \\
GIN                 & $73.38^*$         & $61.25^*$         & $60.33^*$  \\
DGN                 & 79.88         & 65.01         & 75.63  \\
GAT-ECFP            & \underline{90.99} & 67.71    & 75.14  \\
GIN-ECFP            & $82.84^*$         & $64.36^*$         & $54.92^*$  \\
DGN-ECFP            & 89.56         & 64.17         & 73.19  \\
GAT-MFP             & 89.84         & 71.70         & \underline{76.47}  \\
GIN-MFP             & $82.03^*$         & $71.04^*$         & $62.68^*$  \\
DGCL                & \textbf{91.48} & 73.78       & \textbf{81.49 } \\
LGIN Variable $c$(Ours) & 90.30    & \underline{89.08} &  - \\
LGIN Fixed $c$(Ours)     & 86.56    & \textbf{91.54} & 73.27 \\
\bottomrule
\end{tabular}
\end{table}

Table~\ref{tab:molnet} presents performance on the BACE, BBBP, and HIV MoleculeNet datasets, comparing LGIN variants against GAT, GIN, DGN, DGCL, and enhanced versions with ECFP/MFP features. Our method achieves competitive and state-of-the-art results. Specifically, \textbf{LGIN Fixed $c$} attains the highest score on the \textbf{BBBP} dataset (\textbf{91.54}), with \textbf{LGIN Variable $c$} second (\underline{89.08}). On \textbf{BACE}, \textbf{DGCL} leads (\textbf{91.48}), closely followed by \textbf{LGIN Variable $c$} (\underline{90.30}), surpassing GIN-based models (GIN-ECFP 82.84*, GIN-MFP 82.03*). For \textbf{HIV}, \textbf{DGCL} is best (\textbf{81.49}), with GAT-MFP second (\underline{76.47}). LGIN variants consistently outperform baseline GIN models across all datasets. These results demonstrate LGIN's effectiveness in leveraging Lorentzian geometry for molecular representation learning and achieving superior performance in molecular property prediction.

\section{Conclusion and Limitations}
In this paper, we address the need for simple yet effective graph neural networks on hyperbolic manifolds capable of capturing both semantic and structural features. We propose the Lorentzian Graph Isomorphic Network (LGIN), which, to the best of our knowledge, is the first framework to extend graph isomorphism tests to hyperbolic spaces with variable and fixed curvature. However, our analysis does not offer a comprehensive framework for hyperbolic graph neural networks that directly map between hyperbolic spaces without relying on tangent space projections. Further investigation is required to develop such transformations and fully explore the potential of these models, which we leave as a direction for future research.
\bibliography{sample-base}
\bibliographystyle{acm}

\newtheorem*{theorem*}{Theorem}

\appendix
\section{Proof of Theorem \ref{theo:iso}}\label{app:iso}
\begin{proof}
Assume, for contradiction, that such a diffeomorphism \(f : \mathcal{M} \to \mathcal{M}\) exists, satisfying:  
\[
f^* g = g \quad \text{and} \quad f \circ \Phi(G_1) = \Phi(G_2)
\]
Since \(\mathcal{M}\) is a Lorentzian manifold with constant negative curvature \(c < 0\), it admits a unique global isometry group preserving the Lorentzian metric \(g\). Therefore, any isometry \(f\) would map geodesics in \(\mathcal{M}\) to geodesics in \(\mathcal{M}\).  

By definition, the embeddings \(\Phi(G_1)\) and \(\Phi(G_2)\) map vertices of \(G_1\) and \(G_2\) into the hyperboloid model of \(\mathbb{H}^n\), such that graph distances are preserved through the Lorentzian distance function. If \(G_1\) and \(G_2\) are non-isomorphic, then there exists no bijection between their vertex sets preserving adjacency and distances.  

However, the existence of such an \(f\) implies that the embeddings \(\Phi(G_1)\) and \(\Phi(G_2)\) are related by an isometry, contradicting the assumption that \(G_1\) and \(G_2\) are non-isomorphic. Hence, no such \(f\) exists.  

\[
\therefore \Phi(G_1) \neq f \circ \Phi(G_2) \text{ for any isometry } f
\]
\end{proof}

\section{Proof of Proposition \ref{prop:full_rank}}\label{app:full_rank}
\begin{proof}
Let the transformation be denoted by $F(x) = \mathcal{W}(x)$. We can decompose $F$ into a sequence of maps:
$F = f_5 \circ f_4 \circ f_3 \circ f_2 \circ f_1$, where:
\begin{enumerate}
    \item $f_1: H^n \to T_{o_n} H^n$, $f_1(x) = \log_{o_n}^c(x)$. The logarithmic map centered at the origin $o_n$ of the hyperboloid model $H^n$ is a global diffeomorphism from $H^n$ (the upper sheet) to $T_{o_n} H^n$. As a diffeomorphism, $f_1$ is bijective and thus injective.

    \item $f_2: T_{o_n} H^n \to \mathbb{R}^n$, $f_2(\mathbf{v}) = \mathbf{v}_{[1:n]}$. In the hyperboloid model embedded in $\mathbb{R}^{n+1}$, the origin is $o_n=(1, 0, \dots, 0)$. The tangent space at the origin $o_n$ consists of vectors $\mathbf{v} = (v_0, v_1, \dots, v_n) \in \mathbb{R}^{n+1}$ such that $\langle o_n, \mathbf{v} \rangle_\eta = -1 \cdot v_0 + \sum_{i=1}^n 0 \cdot v_i = -v_0 = 0$. Thus $T_{o_n} H^n = \{ (0, v_1, \dots, v_n) \mid v_i \in \mathbb{R} \}$. The map $f_2$ extracts the spatial components $(v_1, \dots, v_n)$, providing an isomorphism between $T_{o_n} H^n$ and $\mathbb{R}^n$. As an isomorphism, $f_2$ is bijective and thus injective.

    \item $f_3: \mathbb{R}^n \to \mathbb{R}^d$, $f_3(\mathbf{y}) = \mathbf{W}\mathbf{y}$. This is a standard linear transformation. A linear transformation $f_3$ is injective if and only if its null space, $\ker(\mathbf{W}) = \{ \mathbf{y} \in \mathbb{R}^n \mid \mathbf{W}\mathbf{y} = \mathbf{0} \}$, contains only the zero vector. By the Rank-Nullity Theorem ($n = \dim(\ker(\mathbf{W})) + \text{rank}(\mathbf{W})$), this is equivalent to $\text{rank}(\mathbf{W}) = n$. The maximum possible rank for $\mathbf{W} \in \mathbb{R}^{d \times n}$ is $\min(d, n)$. Thus, having rank $n$ requires $d \ge n$. $f_3$ is injective if and only if $\mathbf{W}$ has full column rank ($n$).

    \item $f_4: \mathbb{R}^d \to T_{o_d} H^d$, $f_4(\mathbf{z}) = (0, \mathbf{z})$. Similar to $f_2$, this map takes a vector $\mathbf{z} \in \mathbb{R}^d$ and forms a vector $(0, z_1, \dots, z_d) \in \mathbb{R}^{d+1}$, which lies in $T_{o_d} H^d$. This map is an isomorphism between $\mathbb{R}^d$ and $T_{o_d} H^d$. As an isomorphism, $f_4$ is bijective and thus injective.

    \item $f_5: T_{o_d} H^d \to H^d$, $f_5(\mathbf{u}) = \exp_{o_d}^c(\mathbf{u})$. The exponential map centered at the origin $o_d$ is a global diffeomorphism from $T_{o_d} H^d$ to $H^d$ (the upper sheet). As a diffeomorphism, $f_5$ is bijective and thus injective.
\end{enumerate}
The composite function $F = f_5 \circ f_4 \circ f_3 \circ f_2 \circ f_1$ is injective if and only if all functions in the composition are injective. We have shown that $f_1, f_2, f_4,$ and $f_5$ are injective. Therefore, $F$ is injective if and only if $f_3$ is injective.

As shown in point 3, $f_3$ is injective if and only if the matrix $\mathbf{W} \in \mathbb{R}^{d \times n}$ has full column rank ($n$). This implies that the output dimension $d$ must be greater than or equal to the input dimension $n$ ($d \ge n$).

Thus, the Lorentzian transformation defined in eq. \ref{eq:lorentz} is injective if and only if $\mathbf{W}$ has full column rank.

To demonstrate this explicitly, assume $F(x_1) = F(x_2)$ for $x_1, x_2 \in H^n$.
\[ \exp_{o_d}^c \left(0, \mathbf{W} \left(\log_{o_n}^c x_1\right)_{[1:n]}\right) = \exp_{o_d}^c \left(0, \mathbf{W} \left(\log_{o_n}^c x_2\right)_{[1:n]}\right) \]
Since $f_5 = \exp_{o_d}^c$ is injective, we have:
\[ \left(0, \mathbf{W} \left(\log_{o_n}^c x_1\right)_{[1:n]}\right) = \left(0, \mathbf{W} \left(\log_{o_n}^c x_2\right)_{[1:n]}\right) \]
Equating the spatial components (or applying $f_4^{-1}$ which is also injective):
\[ \mathbf{W} \left(\log_{o_n}^c x_1\right)_{[1:n]} = \mathbf{W} \left(\log_{o_n}^c x_2\right)_{[1:n]} \]
Let $\mathbf{y}_1 = \left(\log_{o_n}^c x_1\right)_{[1:n]}$ and $\mathbf{y}_2 = \left(\log_{o_n}^c x_2\right)_{[1:n]}$. Both $\mathbf{y}_1, \mathbf{y}_2 \in \mathbb{R}^n$. The equation is $\mathbf{W}\mathbf{y}_1 = \mathbf{W}\mathbf{y}_2$.
If $\mathbf{W}$ has full column rank, the linear map $f_3(\mathbf{y})=\mathbf{W}\mathbf{y}$ is injective, which implies $\mathbf{y}_1 = \mathbf{y}_2$.
\[ \left(\log_{o_n}^c x_1\right)_{[1:n]} = \left(\log_{o_n}^c x_2\right)_{[1:n]} \]
Since $f_2^{-1}$ maps $\mathbb{R}^n$ back to $T_{o_n} H^n$ injectively, this implies the equality of the tangent vectors:
\[ \log_{o_n}^c x_1 = \log_{o_n}^c x_2 \]
Finally, since $f_1 = \log_{o_n}^c$ is injective, this implies:
\[ x_1 = x_2 \]
Thus, if $\mathbf{W}$ has full column rank, $F(x_1) = F(x_2)$ implies $x_1 = x_2$, proving $F$ is injective.

Conversely, assume $\mathbf{W}$ does not have full column rank. Since $\mathbf{W} \in \mathbb{R}^{d \times n}$, this means $\text{rank}(\mathbf{W}) < n$. By the Rank-Nullity Theorem, $\dim(\ker(\mathbf{W})) = n - \text{rank}(\mathbf{W}) > 0$. This means there exists a non-zero vector $\mathbf{v} \in \mathbb{R}^n$ such that $\mathbf{W}\mathbf{v} = \mathbf{0}$.
Let $\mathbf{u} = f_2^{-1}(\mathbf{v}) \in T_{o_n} H^n$. Since $f_2^{-1}$ is injective and $\mathbf{v} \ne \mathbf{0}$, $\mathbf{u} \ne \mathbf{0}$.
Let $x = f_1^{-1}(\mathbf{u}) \in H^n$. Since $f_1^{-1}$ is injective (it is $\exp_{o_n}^c$) and $\mathbf{u} \ne \mathbf{0}$, $x \ne o_n$.
Consider $F(x)$ and $F(o_n)$.
$F(x) = f_5(f_4(f_3(f_2(f_1(x))))) = f_5(f_4(f_3(\mathbf{u}))) = f_5(f_4(\mathbf{W}\mathbf{v})) = f_5(f_4(\mathbf{0}))$.
$f_4(\mathbf{0}) = (0, \mathbf{0}) \in T_{o_d} H^d$.
$F(x) = \exp_{o_d}^c(0, \mathbf{0})$. The exponential map of the zero vector in the tangent space at $p$ is always $p$. So, $\exp_{o_d}^c(0, \mathbf{0}) = o_d$.
Now consider $F(o_n)$:
$F(o_n) = \exp_{o_d}^c \left(0, \mathbf{W} \left(\log_{o_n}^c o_n\right)_{[1:n]}\right)$.
$\log_{o_n}^c o_n$ is the zero vector in $T_{o_n} H^n$, which is $(0, \mathbf{0}) \in \mathbb{R}^{n+1}$.
$\left(\log_{o_n}^c o_n\right)_{[1:n]} = \mathbf{0} \in \mathbb{R}^n$.
$F(o_n) = \exp_{o_d}^c (0, \mathbf{W}\mathbf{0}) = \exp_{o_d}^c (0, \mathbf{0}) = o_d$.
So $F(x) = o_d$ and $F(o_n) = o_d$. Since $x \ne o_n$, we have found two distinct points $x, o_n \in H^n$ that map to the same point $o_d \in H^d$. Thus, $F$ is not injective if $\mathbf{W}$ does not have full column rank.

Therefore, the Lorentzian transformation $\mathcal{W}$ is injective if and only if the matrix $\mathbf{W}$ has full column rank.
\end{proof}

\section{Preliminaries}
\subsection{Graph Preliminaries}

\textbf{Graph}. Consider the definition of a Graph ${G}=({V},{E})$. Let ${V}$ be the set of vertices $\{v_1,v_2,...v_{n_v}\}$ and $E$ be the set of edges $\{e_1,e_2,.....e_{n_e}\}$. $n_v,n_e$ are the number of vertices and edges in ${G}$ respectively. Each vertex $v$ is characterized by an initial $n$ dimensional vector depending on the problem set. The problem set may be a set of molecular graphs, protein networks, citation networks etc.

\textbf{Weisfeiler-Lehman(WL) Test}. The Weisfeiler-Lehman (WL) test is a graph isomorphism heuristic that iteratively refines node representations based on local neighborhood aggregation. Given a graph, the test assigns initial labels to nodes and updates them iteratively by incorporating labels from neighboring nodes. This process continues until convergence, producing a unique representation for each node that captures its structural role within the graph\citep{morris2019weisfeiler}. The WL test is widely used to distinguish non-isomorphic graphs efficiently and serves as a foundation for many graph neural networks (GNNs)\citep{xu2018powerful}.

\subsection{Definitions of Key Topological Concepts}

\begin{definition}[\textbf{Differential Manifold}]\label{def:3}
 A differential manifold is a topological space $\mathcal{M}$ that is locally homeomorphic to $\mathbb{R}^n$ and consists of a smooth structure that allows for the differentiation of functions. In other words, there is a covering $\{U_i\}$ of $\mathcal{M}$ consisting of open sets $U_i$ homeomorphic to open sets $V_i$ in $\mathbb{R}^n$.
 \end{definition}

\begin{definition}[\textbf{Tangent Space}]\label{def:4}
Let $\mathcal{M}$ be a differential manifold of dimension $n$. We define for every point $p\in M$ a $n$ dimensional vector space $T_p\mathcal{M}$ called the tangent space. The space $T_p\mathcal{M}$ may be defined briefly as the set of all curves $
\gamma: (-a, a) \to M \text{ such that } \gamma(0) = p \text{ and } a > 0$ is arbitrary, considered up to some equivalence relation. The relation is that we identify two curves, that read on some chart $(U_i,\varphi)$ have the same tangent vector at $\varphi_i(p)$. The relationship between the hyperbolic space and the corresponding tangent space is called a map. 
\end{definition}

\begin{definition}[\textbf{Metric Tensor}]\label{def:metric}
A metric tensor on a differentiable manifold $\mathcal{M}$ assigns a smoothly varying inner product to each tangent space $T_p\mathcal{M}$ at every point $p \in \mathcal{M}$. A metric tensor on \( \mathcal{M} \) is a smooth, symmetric, bilinear map  
$$
g: T\mathcal{M} \times T\mathcal{M} \to \mathbb{R}
$$  
such that for each point \( p \in \mathcal{M} \), the restriction \( g_p \) to the tangent space \( T_p\mathcal{M} \) satisfies:  
$$
g_p(X, Y) = g_p(Y, X), \quad \forall X, Y \in T_p\mathcal{M},
$$  
where \( g_p \) is a symmetric, positive-definite bilinear form. This tensor defines an inner product structure on each tangent space, allowing the measurement of angles, lengths, and distances on \( \mathcal{M} \).  Based on this, we can define that the Riemannian manifold is a differentiable manifold with a metric tensor that is positive semi-definite at every point. 
\end{definition}

\begin{definition}[\textbf{Curvature}]
    The curvature of a Riemannian manifold $(\mathcal{M},g)$ is a mathematical entity that measures how distorted the metric tensor $g$ is when compared to the Euclidean structure on $\mathbb{R}^n$. Let \( (\mathcal{M}, g) \) be a Riemannian manifold, where \( g \) is the metric tensor. The Riemann curvature tensor is defined as  
\begin{equation}
    R(X, Y)Z = \nabla_X \nabla_Y Z - \nabla_Y \nabla_X Z - \nabla_{[X,Y]} Z,
\end{equation}
for vector fields \( X, Y, Z \) on \( \mathcal{M} \), where \( \nabla \) is the Levi-Civita connection.  

The sectional curvature at a point \( p \) for a 2-plane \( \sigma \) spanned by \( \{X, Y\} \) in the tangent space \( T_p \mathcal{M} \) is given by  
\begin{equation}
    K(X, Y) = \frac{\langle R(X, Y)Y, X \rangle_g}{\|X\|_g^2 \|Y\|_g^2 - \langle X, Y \rangle_g^2}.
\end{equation}

For a Lorentzian space with constant curvature \( K \), the Riemann tensor satisfies  
\begin{equation}
    R(X, Y)Z = K \left( \langle Y, Z \rangle_g X - \langle X, Z \rangle_g Y \right).
\end{equation}
\end{definition}

\begin{definition}[\textbf{Isometry}]\label{def:5}
A diffeomorphism $f:\mathcal{M}\to \mathcal{N}$ between two Riemannian manifolds $(\mathcal{M},g)$ and $(\mathcal{N},h)$ is an isometry if it preserves the scalar product i.e, the equality,
\begin{equation}
    \langle v,w\rangle=\langle df_p(v),df_p(w)\rangle
\end{equation}
holds for all $p \in \mathcal{M}$ and every pair of vectors $v,w \in T_p\mathcal{M}$. The symbol $\langle,\rangle$ indicates the scalar products in $T_p\mathcal{M}$ and $T_{f(p)}\mathcal{N}$.
\end{definition}

\begin{definition}[\textbf{Parallel Transport}]\label{def:parallel}
Parallel transport is a geometric operation that moves a vector along a curve on a manifold while preserving its inner product with tangent vectors along the path. In the context of hyperbolic geometry, parallel transport ensures that vectors remain tangent to the hyperboloid model while accounting for the manifold's curvature. It is a way to slide frames along geodesics.
\end{definition}

\subsection{Hyperboloid Model}\label{sec:hyperboloid}
In the hyperboloid model, we define $\mathbb{H}^n$ as the set of all points of norm -1 in $\mathbb{R}^{n+1}$, equipped with the Lorentzian scalar product. The Lorentzian scalar product on $\mathbb{R}^{n+1}$ is given by 
\begin{equation}
    \langle x,y\rangle_\eta=-x_0y_0+\sum_{i=1}^{n}x_iy_i
\end{equation}
The Lorentz model of the hyperbolic space, is formally defined as follows:
\begin{equation}
   \mathbb{H}^n = \left\{ x \in \mathbb{R}^{n+1} \mid \langle x, x \rangle_{\eta} = -1, x_0 > 0 \right\}
\end{equation}
The set of points $x$ with $\langle x,x\rangle$ is a hyperboloid with two sheets. As mentioned earlier, maps form the relationship between the hyperbolic space and the corresponding tangent space. Formally, for a point \( p \in\mathcal{ M} \), the exponential map
\[
\exp_p : T_p M \longrightarrow M
\] is defined as follows: A vector \( v \in T_p \mathcal{M} \) determines a maximal geodesic 
\(\gamma_v : \mathbb{R} \to \mathcal{M}\) with \(\gamma_v(0) = p\) and \(\gamma_v'(0) = v\). We set \(\exp_p(v) = \gamma_v(1)\). The logarithmic map $\log_p$ is the inverse of the exponential map. In the hyperboloid space, the exponential and logarithmic maps are given by equations \ref{eq:exp_map} and \ref{eq:log_map} respectively. 
\begin{equation}\label{eq:exp_map}
     \exp_p(v) = \cosh\left(\sqrt{|c|} \|v\|_{\eta} \right) p + \sinh\left(\sqrt{|c|} \|v\|_{\eta} \right) \frac{v}{\|v\|_{\eta}}
\end{equation}
where \( p \in \mathbb{H}^n \), \(c\) is the curvature and \( v \in T_p \mathbb{H}^n \) is a tangent vector at \( p \).

\begin{equation}\label{eq:log_map}
 \log_p(q) = \frac{d_{\mathbb{H}}(p, q)}{\sinh\left(\sqrt{|c|} d_{\mathbb{H}}(p, q) \right)} \left(q - \cosh\left(\sqrt{|c|} d_{\mathbb{H}}(p, q)\right) p \right)
\end{equation}
where 
\begin{equation}
     d_{\mathbb{H}}(p, q) = \frac{1}{\sqrt{|c|}} \cosh^{-1}(-\langle p, q \rangle_{\eta})
\end{equation}
is the hyperbolic distance in the Lorentzian space. The parallel transport is given by:
\begin{equation}
PT^c_{p \rightarrow q} (v) = v - \frac{c\langle q, v \rangle_\eta}{1 + c\langle p, q \rangle_\eta} (p + q)
\end{equation}

\section{Experimental conditions}
\subsection{Datasets}
For our experiments, we use 6 Bioinformatics datasets and 3 MoleculeNet datasets. The bioinformatics datasets include, namely, Proteins, Mutag, NCI1, PTC, Enzymes, and DD\citep{yanardag2015deep}. The MoleculeNet datasets include, namely, BBBP, BACE and HIV\citep{wu2018moleculenet}. The initial features of the Bioinformatics datasets are mostly the node degree, while for MoleculeNet, each node is assigned 9 features such as atomic number, chirality, formal charge, or whether the atom is in a ring or not. Datasets have been described in detail in appendix \ref{app:dataset}. 

\subsection{Hyperparameters and model configuration}
In our proposed method, we evaluate on sets of variants. One with a constant curvature $c$ and the other with variable curvature, which is essentially a trainable parameter. All initial curvatures were set to 4. The trainable parameter $\epsilon$ was initially set to 0.1. We use the Riemannian Adam\citep{becigneul2018riemannian} to train our model with a cosine annealing learning rate scheduler with warm restarts. We use the same model architecture for all datasets, which is a 3-layer Lorentzian Graph Isomorphic Network predicting 128, 256 and 512 features respectively. To ensure stability of gradient flow, we make use of gradient clipping after gradient computation at each step. The number of Lorentz Transformation layers has been set to 2. 

\subsection{Evaluation Protocol}
 Unless otherwise mentioned, for Bioinformatics datasets, we record the mean and standard deviation of the test accuracy on 10 splits. For MoleculeNet datasets, we consider the average accuracy on 3 splits.

\section{Ablations on Curvature and Parallel Transport}

\begin{figure*}
    \centering
    \includegraphics[scale=0.8]{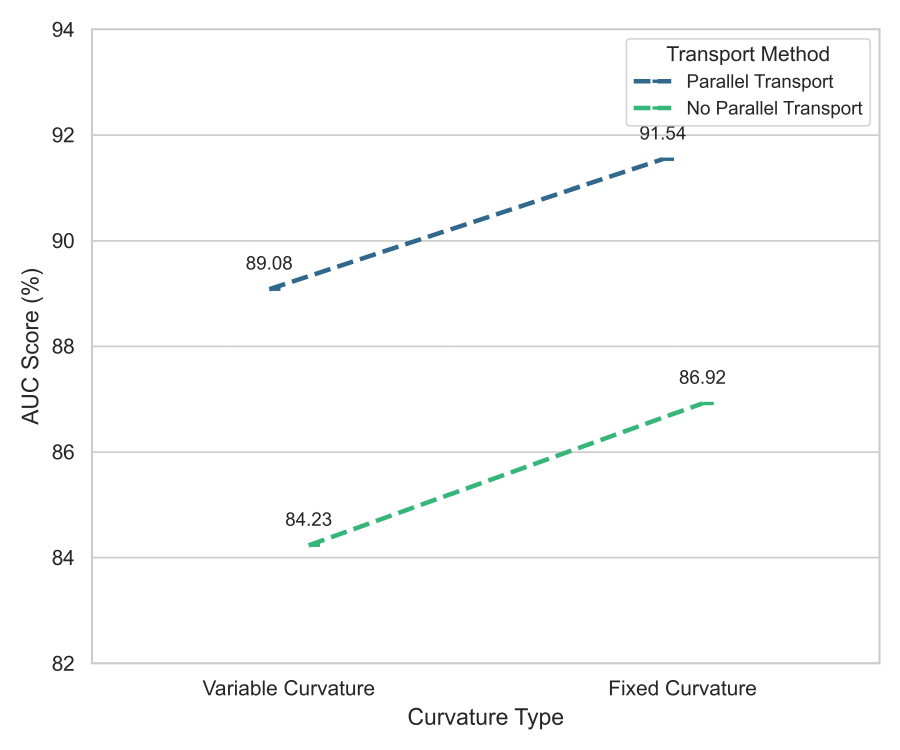}
    \caption{Abalation Study on Curvature Type  and Parallel Transport on BBBP}
    \label{fig:line}
\end{figure*}

In this section, we provide an ablation study on parallel transport and curvatures on the BBBP dataset. As mentioned earlier, in eq.\ref{eq:update}, we use parallel transport between 2 tangent spaces to ensure geometric consistency and numerical stability. We perform an analysis by considering the following equation i.e eq.\ref{eq:ablation} as well, and show the importance of parallel transport by measuring the performance difference. 
 \begin{equation}
\mathbf{x}^{\mathcal{L}^k}=\text{LT}^k(\exp_o^c((1+\epsilon^k).\log_o^c(\mathbf{x}^{\mathcal{L}^{k-1}})+\log_o^c(T(\mathbf{x}^{\mathcal{L}^{k-1}}))))
     \label{eq:ablation}
 \end{equation}

Figure \ref{fig:line} represents the AUC scores for curvature and parallel transport modes while  provides a direct comparison between the results of the two update rules. As noted, cases where parallel transport is employed to the first term of eq.\ref{eq:ablation} perform considerably better. We hypothesize that this is probably due to the fact that parallel transport aligns embeddings which were otherwise distorted due to aggregation, despite all transformations taking place with respect to the origin. With respect to curvature type, there is no significant difference between fixed and variable curvature models, in line with evaluations described in previous sections and \citep{zhang2021lorentzian}.

\section{Detailed Related Work}\label{app:related}
We detail the related work discussed in the main body of this paper here. 
\subsection{Advances in Hyperbolic Graph Neural Networks}
Hyperbolic graph neural networks, as discussed above, have emerged as a powerful method for modeling complex and hierarchical structures. \citep{liu2019hyperbolic} generalized graph neural networks to be manifold-agnostic and demonstrated that hyperbolic graph neural networks are more efficient at capturing structural properties than their Euclidean counterpart. \citep{khatir2022unification} proposed the Poincare disk model as the search space to apply all approximations on the disk, thus eliminating the need for inter-space transformations. This framework introduces a hyperbolic normalization layer that simplifies the entire hyperbolic model to a Euclidean model cascaded with the normalization layer, maintaining the advantages of both geometric approaches. \citep{fu2023adaptive} introduced ACE-HGNN which dynamically learns the optimal curvature based on the input graph and downstream tasks. Using a multi-agent reinforcement learning framework with two agents—ACE-Agent and HGNN-Agent—for learning curvature and node representations respectively, this model adapts to the specific hierarchical structures present in different graphs rather than using a manually fixed curvature value. \citep{zhang2021lorentzian} introduced Lorentzian Graph Convolutional Networks(LGCN) for modeling hierarchical architectures. They design a neighborhood aggregation method based on the centroid of Lorentzian distance to constrain embeddings within the hyperboloid. Further, they theoretically prove that some proposed graph operations are equivalent to those defined in other hyperbolic models such as the Poincare ball model. \citep{choudhary2023hyperbolic} introduced H-GRAM, a meta-learning framework specifically designed for hyperbolic graph neural networks. The model learns transferable information from local subgraphs in the form of hyperbolic meta gradients and label hyperbolic protonets thereby enabling faster learning over new tasks dealing with disjoint subgraphs, thus addressing the generalization challenge.

\subsection{Applications of Hyperbolic Graph Neural Networks}
\citep{ramirez2025fully} used a Fully Hyperbolic Graph Neural Network(FHGNN) to embed functional brain connectivity graphs derived from magnetoencephalography (MEG) data into low dimensions on a Lorentz model of hyperbolic space. \citep{wu2021hyperbolic} developed a graph-based Quantitative Structure-Activity Relationship(QSAR) method by building a hyperbolic relational graph convolution network. It leverages both molecular structure and molecular descriptors to achieve state-of-the-art performance on 11 drug discovery-related datasets.  \citep{bao2025hgcge} introduced Hyperbolic Graph Convolutional Network-based Knowledge Graph Embedding (HGCGE) which addresses challenges in traditional GCN-based KGE methods, such as oversmoothing and high distortion in Euclidean space. It employs GCN operations in hyperbolic space with Möbius transformations to embed entities and relationships in the Poincaré model, enhancing hierarchical data representation. The proposed scoring function improves entity distinction across relationships, achieving superior performance on multiple datasets, even at low dimensions and training rounds.

\subsection{Graph Expressiveness}
\citep{xu2018powerful} presented a theoretical framework for analyzing the expressive power of GNNs. Their analysis demonstrated that popular GNN variants such as Graph Convolutional Networks and GraphSAGE cannot distinguish certain simple graph structures, limiting their discriminative capabilities. The researchers then developed a simple architecture—the Graph Isomorphism Network—that is provably the most expressive among the class of message-passing GNNs and as powerful as the 1-dimensional Weisfeiler-Lehman(WL) graph isomorphism test. \citep{hevapathige2023uplifting} showed that partitioning a graph into sub-graphs that preserve structural properties provides a powerful means to exploit interactions among different structural components of the graph. The researchers proposed Graph Partitioning Neural Networks(GPNNs) that combines structural interactions via permutation invariant graph partitioning to enhance graph representation learning.  \citep{zhang2020improving} showed that attention-based GNNs may face limitations in discriminative power due to non-injective aggregation functions. This non-injectivity can lead to different substructures being mapped to the same representation, reducing the model's effectiveness. To mitigate this, methods have been proposed to enhance injectivity, such as preserving the cardinality information of multisets during aggregation. \citep{lim2024contextualized} introduced the concept of soft-injective functions. These functions aim to approximate injectivity by ensuring that distinct inputs are mapped to sufficiently different outputs, considering a predefined dissimilarity measure. This approach allows GNNs to maintain discriminative power without necessitating strictly injective aggregation functions.

\section{Reproducibility}
In this section, we describe the datasets and hyperparameters used
\subsection{Code and Dataset Description}\label{app:dataset}
The code associated with our paper can be found at \hyperref[https://github.com/Deceptrax123/LGIN]{https://github.com/Deceptrax123/LGIN}. All the data used in our study are from open source repositories and can be found at https://pytorch-geometric.readthedocs.io/en/latest/modules/datasets.html. We describe the datasets we used in our study below. 
\begin{enumerate}
    \item \textbf{MUTAG}: A collection of nitroaromatic compounds and the goal is to predict their mutagenicity on Salmonella typhimurium. Input graphs are used to represent chemical compounds, where vertices stand for atoms and are labeled by the atom type (represented by one-hot encoding), while edges between vertices represent bonds between the corresponding atoms. It includes 188 samples of chemical compounds with 7 discrete node labels
    \item \textbf{PROTEINS}: A dataset of proteins that are classified as enzymes or non-enzymes. Nodes represent the amino acids and two nodes are connected by an edge if they are less than 6 Angstroms apart
    \item \textbf{PTC}: A collection of 344 chemical compounds represented as graphs which report the carcinogenicity for rats. There are 19 node labels for each node.
    \item \textbf{NCI1}: The NCI1 dataset comes from the cheminformatics domain, where each input graph is used as representation of a chemical compound: each vertex stands for an atom of the molecule, and edges between vertices represent bonds between atoms. This dataset is relative to anti-cancer screens where the chemicals are assessed as positive or negative to cell lung cancer. Each vertex has an input label representing the corresponding atom type, encoded by a one-hot-encoding scheme into a vector of 0/1 elements.
    \item \textbf{DD}: A widely used graph classification benchmark derived from the {Protein Data Bank (PDB)}. It consists of {1178} graphs, where each graph represents a protein structure. Nodes typically correspond to amino acid residues, and edges connect residues based on spatial proximity (e.g., within a threshold distance between C$\alpha$ atoms). The task on DD is {binary graph classification}, requiring models to distinguish between two structural classes of proteins. 
    \item \textbf{ENZYMES}: A dataset of 600 protein tertiary structures obtained from the BRENDA enzyme database. The ENZYMES dataset contains 6 enzymes.
    \item \textbf{HIV}: The HIV dataset was introduced by the Drug Therapeutics Program (DTP) AIDS Antiviral Screen, which tested the ability to inhibit HIV replication for over 40,000 compounds. Screening results were evaluated and placed into three categories: confirmed inactive (CI), confirmed active (CA), and confirmed moderately active (CM)
    \item \textbf{BBBP}: The BBBP dataset comes from a study focused on modeling and predicting the permeability of the blood-brain barrier. The BBBP dataset contains binary labels indicating whether a compound can penetrate the blood-brain barrier (BBB) or not. Researchers use this dataset to develop and evaluate machine learning methods for predicting BBB permeability. It’s a critical task because understanding which compounds can cross the BBB is essential for drug discovery and designing therapeutics for neurological conditions.
    \item \textbf{BACE}: The BACE dataset focuses on inhibitors of human beta-secretase 1 (BACE-1). It includes both quantitative (IC50 values) and qualitative (binary labels) binding results. The dataset comprises small molecule inhibitors across a wide range of affinities, spanning three orders of magnitude (from nanomolar to micromolar IC50 values). Specifically, it provides: 154 BACE inhibitors for affinity prediction. 20 BACE inhibitors for pose prediction. 34 BACE inhibitors for free energy prediction.
\end{enumerate}

\subsection{Hyperparameters}
We state the hyperparameters used for most experiments in table \ref{tab:hyperparameters}. 

\begin{table*}[htbp]
  \centering
  \caption{Hyperparameters used for LGIN} 
  \label{tab:hyperparameters}
  \begin{tabular}{lc}
    \toprule
    Parameter & Value \\
    \midrule
    \multicolumn{2}{l}{\textit{Optimizer Parameters}} \\ 
    Learning Rate (LR) & $10^{-3}$ \\
    Adam Betas & (0.9, 0.999) \\ 
    Adam Epsilon & $10^{-8}$ \\
    Gradient Clip Max Norm & 1.0 \\
    \midrule
    \multicolumn{2}{l}{\textit{Model Architecture Parameters}} \\ 
    Numerical Stability Epsilon & 0.1 \\ 
    Number of MLP Layers & 2 \\
    Initial Curvature ($C_{in}$) & 4.0 \\
    Output Curvature ($C_{out}$) & 4.0 \\
    Use Attention & True \\
    Use Bias & True \\
    Dropout Rate & 0-0.2 \\
    \bottomrule
  \end{tabular}
\end{table*}

\section{Ackowledgement}
We would like to thank Vellore Institute of Technology for supporting us throughout this research.

\end{document}